\documentclass[letterpaper, 10 pt, conference]{ieeeconf}  

\IEEEoverridecommandlockouts                              

\usepackage[utf8]{inputenc}
\usepackage[T1]{fontenc}
\usepackage{graphicx}
\usepackage{multicol}
\usepackage{dblfloatfix}
\usepackage{epsfig} 
\usepackage{amsmath} 
\usepackage{amssymb}  
\usepackage{subfig}
\usepackage{float}
\usepackage{hyperref}

\usepackage{color}

\usepackage[sc]{mathpazo}
\usepackage[T1]{fontenc}

\usepackage[font=footnotesize,labelfont=bf]{caption}
\usepackage{cite}
\renewcommand{\baselinestretch}{1}
\newcommand{\R}{\mathcal{R}}
\newcommand{\Hu}{\mathcal{H}}

\title{\LARGE \bf
	Courteous Autonomous Cars
}

\author{Liting Sun$^{1}$, Wei Zhan$^{1}$, Masayoshi Tomizuka$^{1}$, and Anca D. Dragan$^{2}$
	\thanks{$^{1}$Liting Sun, Wei Zhan and Masayoshi Tomizuka are with the Department of Mechanical Engineering, University of California, Berkeley, CA, USA, 94720.  {\tt\small \{litingsun, wzhan, tomizuka\}@berkeley.edu}}
	\thanks{$^{2}$Anca D. Dragan is with the Department of Electrical Engineering and Computer Sciences, University of California, Berkeley, CA, USA, 94720. {\tt\small anca@berkeley.edu}}
}

\begin{document}
	\maketitle
	\thispagestyle{empty}
	\pagestyle{empty}

\begin{abstract}
Typically, autonomous cars optimize for a combination of safety, efficiency, and driving quality. But as we get better at this optimization, we start seeing behavior go from too conservative to too aggressive. The car's behavior exposes the incentives we provide in its cost function. In this work, we argue for cars that are not optimizing a purely selfish cost, but also try to be \emph{courteous} to other interactive drivers. We formalize courtesy as a term in the objective that measures the increase in another driver's cost induced by the autonomous car's behavior. Such a courtesy term enables the robot car to be aware of possible irrationality of the human behavior, and plan accordingly.
We analyze the effect of courtesy in a variety of scenarios. We find, for example, that courteous robot cars leave more space when merging in front of a human driver. 
 Moreover, we find that such a courtesy term can help explain real human driver behavior on the NGSIM dataset.
\end{abstract}

\section{Introduction}

Autonomous cars are getting better at generating their motion not only in isolation, but also around people. We now have many strategies for dealing with interactions with people on the road, each modeling people in substantially different ways. 

Most techniques first anticipate what people plan on doing, and generate the car's motion to be efficient, but also to safely stay out of their way. This prediction can be as simple as assuming the person will maintain their current velocity within the planning horizon \cite{kuwata2009real,liang2012automatic,zhan_spatially-partitioned_2017}, or as complicated as learning a human driver policy or cost function \cite{alahi2016social,zhan_non-conservatively_2016,shimosaka2015predicting,Levine2012ICML}. 

Other techniques account for the interactive nature of coordinating on the road, and model people as changing their plans depending on what the car does. Some do it via coupled planning, assuming that the person and the robot are on the same team, optimizing the same joint cost function \cite{de2013autonomous,hafner2011automated,kretzschmar2016socially}, while others capture interaction as a game in which the human and robot have different utilities, but they influence each other's actions \cite{sadigh2016planning,bahram_game-theoretic_2016,li_game_2017}. 

All of these works focus on \emph{how} to optimize the robot's cost when the robot needs to interact with people. In this paper, we focus on \emph{what} the robot should optimize in such situations, particularly if we consider the fact that humans are not perfectly rational.

Typically, when designing the robot's cost function, we focus on safety and driving quality of the ego vehicle. Arguably, that is rather \emph{selfish}. 

Selfishness has not been a problem with approaches that predict human plans and react to them, because that led to conservative robots that always try to stay out of the way and let people do what they want. But, as we are switching to more recent approaches that draw on the game-theoretic aspects of interaction, our cars are starting to become more aggressive. They cut people off, or inch forward at intersections to go first~\cite{sadigh2016planning}\cite{sadighiros2016}. While this behavior is good sometimes, we would not want to see it all the time. 

Our observation is that as we get better at solving the optimization problem for driving by better models of the world and of the people in it, there is an increased burden on the cost function we optimize to capture what we want. We propose that purely selfish robots that care about their safety and driving quality are not good enough. They should also be \emph{courteous} to other drivers. This is of crucial importance since humans are not perfectly rational, and their behavior will be influenced by the aggressiveness of the robot cars.


\emph{We advocate that a robot should balance minimizing the inconvenience it brings to another driver, and that we can formalize inconvenience as the increase in the other driver's cost due to the robot's behavior to capture one aspect of human behavior irrationality.}

We make the following contributions:

\noindent\textbf{A formalism for courtesy incorporating irrational human behavior.} We formalize courteous planning as trading off between the robot's selfish objective and a courtesy term, and introduce a mathematical definition for this term for irrational human behavior \--- we measure the increase of the vehicle's best cost under the robot's planned behavior, compared to the vehicle's best cost under an alternative "best case scenario", and define the cost increase as the courtesy term.

\noindent\textbf{An analysis of the effects of courteous planning.} We show the difference between courteous and selfish robots under different traffic scenarios. The courteous robot leaves the person more space when it merges, and might even block another agent (not a person) to ensure that the human can safely proceed.

\noindent\textbf{Showing that courtesy helps explain human driving.} We do an Inverse Reinforcement Learning (IRL)-based analysis \cite{Levine2012ICML, abbeel2004apprenticeship, ziebart2008maximum, abbeel2011inverse} to study whether our courtesy term helps in better predicting how humans drive. On the NGSIM dataset \cite{alexiadis_next_2004} of real human driver trajectories, we find that courtesy produces trajectories that are significantly closer to the ground truth.

We think that the autonomous car of the future should be safe, efficient, and courteous to others, perhaps even more so than represented in our current human-only driving society. Our paper enables autonomous car designers to decide to make that happen.

\section{Problem Statement} 
\label{sec:problem_formulation}
In this paper, we consider an interactive robot-human system with two agents: an autonomous car $\mathcal{R}$ and a human driver $\mathcal{H}$ \footnote{If there are multiple robot cars that we control, we treat them all as a single $\mathcal{R}$. If there are multiple human drivers, we reason about how each of them affects the robot's utility separately.}. Our task is to enable a courteous robot car which cares about the potential inconvenience it brings to the human driver's utilities, and generates trajectories that are socially predictable and acceptable.

Throughout the paper, we denote all robot-related terms by subscript $(\cdot)_{\mathcal{R}}$ and all human-related terms by $(\cdot)_\mathcal{H}$.

Let $x_{\mathcal{R}}$ and $u_{\mathcal{R}}$ denote, respectively, the robot's state and control input, and $x_{\mathcal{H}}$ and $u_{\mathcal{H}}$ for the human's. $x{=}(x_{\mathcal{R}}^T,x_{\mathcal{H}}^T)^T$ represents the states of the interaction system. For each agent, we have 
\begin{eqnarray}
x^{t{+}1}_{\mathcal{R}}=f_{\mathcal{R}}\left(x^t_{\mathcal{R}}, u^t_{\mathcal{R}}\right),\\
x^{t{+}1}_{\mathcal{H}}=f_{\mathcal{H}}\left(x^t_{\mathcal{H}}, u^t_{\mathcal{H}}\right),
\end{eqnarray}
and the overall system dynamics are
\begin{eqnarray}
x^{t{+}1}=f\left(x^t, u^t_{\mathcal{R}}, u^t_{\mathcal{H}}\right).\label{eq:open-loop-dynamics}
\end{eqnarray}

We assume that both the human driver and the autonomous car are optimal planners, and they use Model Predictive Control (MPC) with a horizon of length $N$. Let $C_{\mathcal{R}}$ and $C_{\mathcal{H}}$ be, respectively, the cost functions of the robot car and the human driver over the horizon: 
\begin{equation}
C_{i}\left(x^t, \mathbf{u}_{\mathcal{R}}, \mathbf{u}_{\mathcal{H}}; \theta_{i}\right) =\sum_{k=0}^{N{-}1} c_{i}\left(x^{t,k}, u^k_{\mathcal{R}}, u^k_{\mathcal{H}}; \theta_{i}\right), i{\in}\{\R,\Hu\}
\end{equation}
where $\mathbf{u}_{i}{=}(u^0_{i}, u^1_{i}, \cdots, u^{N{-}1}_{i})^T$ are sequences of control actions of the robot car ($i{=}\R$) and the human driver ($i{=}\Hu$), and $x^{t,k}$ with $k{=}0,1,\cdots,N{-}1$ are the corresponding sequence of system states. $\theta_i$ represent, respectively, the preferences of the robot car ($i{=}\R$) and the human driver ($i{=}\Hu$). At every time step $t$, the robot car and the human driver generate their optimal sequences of actions $\mathbf{u}^*_{\mathcal{R}}$ and $\mathbf{u}^*_{\mathcal{H}}$ by minimizing $C_{\mathcal{R}}$ and $C_{\mathcal{H}}$, respectively, execute the first steps $u^{*0}_{\mathcal{R}}$ and $u^{*0}_{\mathcal{H}}$ (i.e., set $u^t_{i}{=}u^{*0}_{i}$ in (\ref{eq:open-loop-dynamics})), and replan for step $t{+}1$. 

Such an optimization-based state feedback strategy formulates the closed-loop dynamics of the robot-human interaction system as a game. To simplify the game, we assume that the robot car has access to $C_{\mathcal{H}}$, and that the human only computes a best response to the robot's actions rather than trying to influence them, as in \cite{sadigh2016planning}. This means that the robot car can compute, for any control sequence it considers, how the human would respond and what cost the human will incur:
\begin{IEEEeqnarray}{rCl}
&&\mathbf{u}^{*}_{\mathcal{H}}=\arg\min_{\mathbf{u}_{\mathcal{H}}} C_{\mathcal{H}}\left(x^t,  \mathbf{u}_{\mathcal{R}}, \mathbf{u}_{\mathcal{H}}; \theta_{\mathcal{H}}\right)\triangleq g(x^t, \mathbf{u}_{\mathcal{R}}; \theta_{\mathcal{H}})\quad\label{eq:response_curve}\\
&&C^*_{\mathcal{H}}(\mathbf{u}_{\mathcal{R}})=C_{\mathcal{H}}\left(x^t, \mathbf{u}_{\mathcal{R}}, g(x^t, \mathbf{u}_{\mathcal{R}}; \theta_{\mathcal{H}}); \theta_{\mathcal{H}}\right)\label{eq:response_cost}.
\end{IEEEeqnarray}
Here $g(x^t, \mathbf{u}_{\mathcal{R}}; \theta_{\mathcal{H}})$ represents the response curve of the human driver towards the autonomous car.

Armed with this model, the robot can now compute what it should do, such that when the human responds, the combination is good for the robot's cost:
\begin{equation}
  \mathbf{u}^*_{\R}=\arg\min_{\mathbf{u}_{\R}} C_{\R}\left(x^t, \mathbf{u}_{\R}, g(x^t, \mathbf{u}_{\R}; \theta_{\mathcal{H}});    \theta_{\R}\right)\label{eq:robot_opt}.
\end{equation}

Our goal is to generate courteous robot behavior to the human, i.e. that takes into consideration the inconvenience it brings to the human driver. We will do so by changing the cost function of the robot to reflect this inconvenience.

\section{Courteous Planning}
\label{sec:courtesy_modeling}
We propose a \emph{courteous planning} strategy based on one key observation: human is not perfectly rational, and one of the irrationality is that they weight losses higher than gains when evaluating their actions \cite{tversky1992advances}. Hence, a courteous robot car should balance the minimization of its own cost function and the inconvenience (loss) it brings to the human driver.

Therefore, we construct $C_{\R}$ in (\ref{eq:robot_opt}) as
\begin{IEEEeqnarray}{rCl}
C_{\R}\left(x^t{,}\mathbf{u}_{\R},\mathbf{u}_{\Hu}{;}\theta_{\R}{,}\theta_{\Hu}{,}\lambda_c\right) &{=}&C^{self}_{\R}\left(x^t{,}\mathbf{u}_{\R}{,}\mathbf{u}_{\mathcal{H}}{;}\theta_{\R}\right)\nonumber\\
+\lambda_c &&C^{court}_{\R}\left(x^t,\mathbf{u}_{\R},\mathbf{u_{\mathcal{H}}};\theta_{\mathcal{H}}\right),\label{eq:courtesy_cost}
\end{IEEEeqnarray}
where $C^{self}_{\R}$ is the cost function for a regular (selfish) robot car which cares about only its own utilities (safety, efficiency, etc), and $C^{court}_{\R}$ models the courtesy term of the robot car to the human driver. It is a function of the robot car's behavior, the human's behavior, the human's cost parameters ($\theta_{\Hu}$) and some alternative costs (see Section \ref{sec:courtesy_modeling}.A). $\lambda_c{\in}[0,\infty)$ captures the trade-off. If we want the robot car to be just as courteous as a human driver, we could learn $\lambda_c$ from human driver demonstration, as we do in Section \ref{sec:IRL}. As robot designers, we might set this parameter higher than regular human driving to enable more courteous autonomous cars, particularly when they do not have passengers on board.



\subsection{Alternative Costs}
With any robot plan $\mathbf{u}_{\R}$, the robot car changes the human driver's environment and therefore induces a best cost for the human, $C_{\Hu}^*(\mathbf{u}_{\R})$. Our courtesy term compares this cost with the \emph{alternative}, $C_{\Hu}^{alt,*}$ \--- the best case scenario for the person. It is not immediately clear how to define this best case scenario since it may vary depending different on driving scenarios. We explore three alternatives.

\vspace{1em}
\noindent\textbf{What the human could have done, had the robot car not been there.}
We first consider a world in which the robot car wouldn't even exist to interfere the person. In such a world, the person gets to optimize their cost without the robot car:
\begin{equation}
C_{\Hu}^{alt,*}(x^t,\theta_{\Hu})=\min_{\mathbf{u}_{\Hu}}C_{\Hu}(x^t,\mathbf{u}_{\Hu};\theta_{\Hu})
\end{equation}
This induces a very generous definition of courtesy: the alternative is for the robot car to not have been on the road at all. In reality though, the robot car is there, which leads to our second alternative.

\vspace{0.2em}
\noindent\textbf{What the human could have done, had the robot car only been there to help the human.}
Our second alternative is to assume that the robot car already on the road could be completely altruistic. The robot car could actually optimize the human driver's cost, being a perfect collaborator:
\begin{equation}
C_{\Hu}^{alt,*}(x^t,\theta_{\Hu})=\min_{\mathbf{u}_{\Hu},\mathbf{u}_{R}}C_{\Hu}(x^t,\mathbf{u}_{\R},\mathbf{u}_{\Hu};\theta_{\Hu})
\end{equation}
For this alternative, the robot car and the human would perform a joint optimization for the human's cost. For example, the robot car can brake to make sure that the human could change lanes in front of it, or even block another traffic participant to make sure the human has space.

\vspace{0.2em}
\noindent\textbf{What the human could have done, had the robot car just kept doing what it was previously doing.}
A fully collaborative robot car is still perhaps not the fairest one to compute inconvenience against. After all, the autonomous car does have a passenger sometimes, and it is fair to take their needs into account too. Our third alternative computes how well the human driver could have done, had the robot car kept acting the same way as it was previously doing:
\begin{equation}
C_{\Hu}^{alt,*}(x^t,\theta_{\Hu})=\min_{\mathbf{u}_{\Hu}}C_{\Hu}(x^t,\mathbf{u}_{\R}^{t-1},\mathbf{u}_{\Hu};\theta_{\Hu})
\end{equation}
This means that the person is now responding to a constant robot trajectory $\mathbf{u}_{\R}^{t-1}{=}(u_{\R}^{t-1},..,u_{\R}^{t-1})$, for instance, maintaining its current velocity. 

Our experiments below explore these three different alternative options for the courtesy term.

\subsection{Courtesy Term}
We define the courtesy term based on the difference between what cost the human has, and what cost they would have had in the alternative:

\noindent\textbf{Definition 1} (Courtesy of the Robot Car)
\begin{IEEEeqnarray}{rCl}
C_{\R}^{court}(x^t,\mathbf{u}_{\R},\mathbf{u}_{\mathcal{H}};\theta_{\mathcal{H}})&=&\max\{0, C_{\mathcal{H}}(x^t,\mathbf{u}_{\R},\mathbf{u}_{\mathcal{H}};\theta_{\mathcal{H}})\nonumber\\ &&\qquad\qquad- C_{\mathcal{H}}^{alt,*}(x^t;\theta_{\mathcal{H}})\}\label{eq:courtesy_def}
\end{IEEEeqnarray}

Note that we could have also sent the courtesy term to simply be the human cost, and have the robot trade off between its cost and the human's. However, that would have penalized the robot for any cost the human incurs, even if the robot does not bring any inconvenience to the human. That might cause too conservative behavior. In fact, if we treat the alternative cost as the reference point in Prospect Theory \--- a human irrationality model \cite{tversky1992advances}, then the theory suggests that human weigh losses more than gains. This means that our courteous robot car should care more about avoiding additional inconvenience, rather than providing more convenience, i.e., helping to reduce the human cost lower than the alternative one. Mathematically, this concept is formulated via Definition 1: the robot does not get any bonus for bringing the human cost lower than $C_{\Hu}^{alt,*}$ (possible with some definitions of $C_{\Hu}^{alt,*}$), it only gets a penalty for making it higher. 

\subsection{Solution}
Thus far, we have constructed a compound cost function $C_{\R}(x^t,\mathbf{u}_{\R}, \mathbf{u}_{\Hu};\theta_{\R}, \theta_{\Hu},\lambda_c)$ to enable a courteous robot car, considering three alternative costs. At every step, the robot needs to solve the optimization problem in (\ref{eq:robot_opt}) to find the best actions to take. We approximate the solution by alternatively fixing one of $\mathbf{u}_{\R}$ or $\mathbf{u}_{\Hu}$, and solving for the other.


\section{Analysis of Courteous Planning}
\label{sec:case_study}

In this section, we analyze the effect of courteous planning on the robot's behavior in different simulated driving scenarios. In Section \ref{sec:IRL}, we study how courteous planning can help better explain real human driving data, enabling robots to be more human-like and predictable, as well as better able at anticipating human driver actions on the road.

\noindent\textit{Simulation Environment}: We implement the simulation environment using Julia \cite{julia} on a 2.5 GHz Intel Core i7 processor with 16 GB RAM. We set the horizon length to $N{=}10$, and the sampling time to 0.1s. Our simulated environment is 1/10 scale of the real world: 1/10 road width, car sizes, maximum acceleration (0.5$m/s^2$) and deceleration (-1.0$m/s^2$), and low speed limit (1.0m/s). 

Regarding the cost functions $C_{\Hu}$ and $C_{\R}$ in (\ref{eq:response_cost})-(\ref{eq:courtesy_cost}),  except for the courtesy term formulated above, we penalize safety, car speed, comfort level and goal distances in both $C_{\Hu}$ and $C^{self}_{\R}$. Details about this can be found later in Section \ref{sec:IRL}.

For all results, we denote a selfish (baseline) autonomous car with gray rectangle, a courteous one as orange, and the human driver as dark blue.

\subsection{The Effect of Courtesy}
\begin{figure*}[t!]
	\begin{centering}
		\includegraphics[scale=0.58]{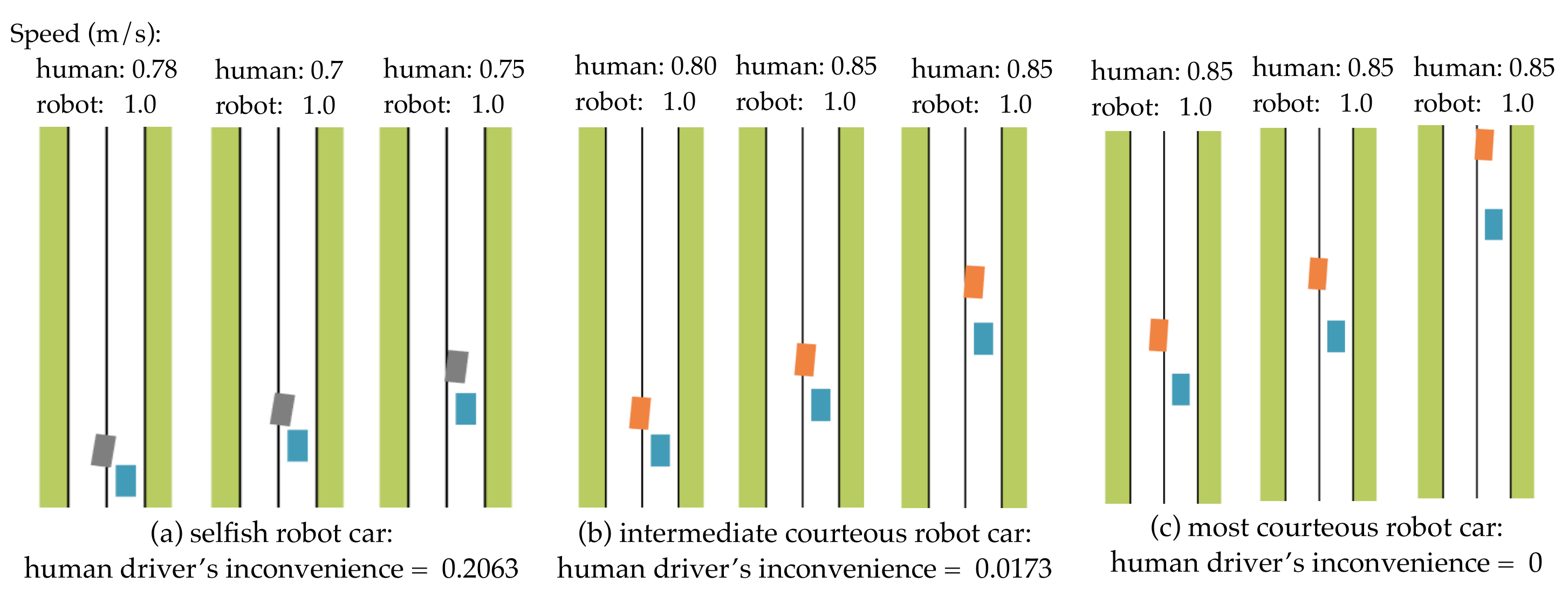}
		\par\end{centering}
	\caption{A lane changing scenario: both the human car and robot car speed at 0.85 m/s initially;  (a) a selfish robot car merges in front of the human with a small gap so that the human brakes to yield; (b) an intermediate courteous robot car merges with a larger gap, which releases the human driver from hard brakes; (c) a most courteous robot car merges with a gap large enough so that the human can maintain speed.\label{fig:lane_change_low_speed_Results}}
\end{figure*}
\begin{figure}[ht!]
	\begin{centering}
		\includegraphics[scale=0.58]{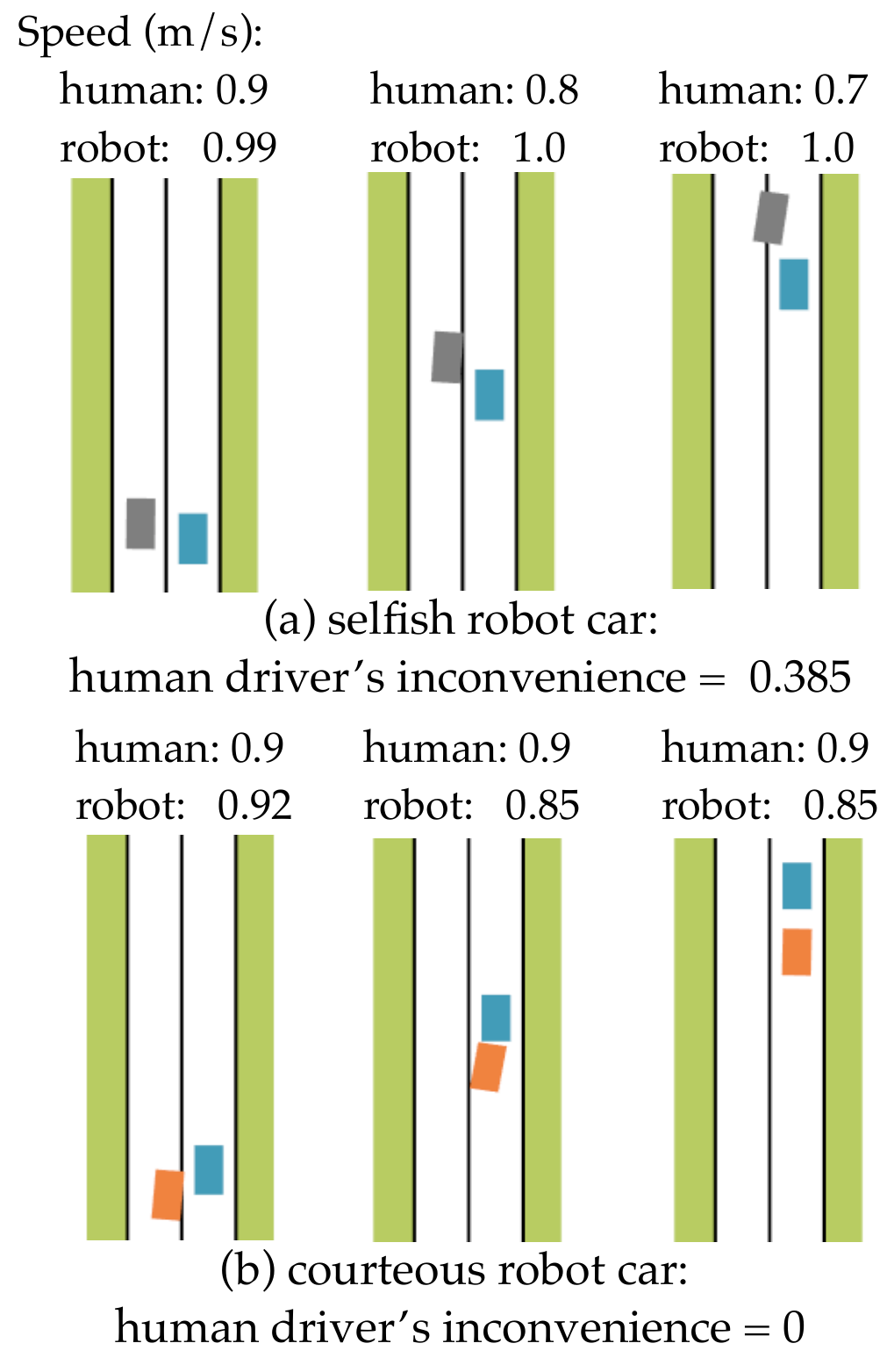}
		\par\end{centering}
	\caption{Another lane changing scenario: both the human car and robot car speed at 0.9 m/s initially; (a) a selfish robot car accelerates and merges in front of the human driver with a small gap, scaring the human driver to brake; (b) a courteous robot car decelerates and merges after the human driver so that the human can maintain speed. \label{fig:lane_change_fast_speed_Results}}
\end{figure}

\subsubsection{Lane Changing}
We first consider a lane changing driving scenario, as shown in Fig.~\ref{fig:lane_change_low_speed_Results}. The autonomous car wants to merge into the human driver's lane from an adjacent lane. We assume that the goal of the human driver is to maintain speed. Then all three different alternatives lead to the same alternative optimal behavior and cost of the human: the human would go in their lane undisturbed by the robot. Hence, with constant $C_{\Hu}^{alt,*}$, we focus on the influence of the trade-of factor $\lambda_c$ in the results.

We present two sets of simulation results in Fig.~\ref{fig:lane_change_low_speed_Results} and Fig.~\ref{fig:lane_change_fast_speed_Results}, where the initial human driver's speeds are 0.85 m/s and 0.9 m/s respectively. The results show that as $\lambda_c$ increases, i.e., being more courteous, the autonomous car tends to leave a larger gap when it merges in front of the human, and the human brakes less (Fig.~\ref{fig:lane_change_low_speed_Results} from left to right). When the human driver's initial speed is high enough, a courteous autonomous car decides to merge afterwards instead of cutting in, as shown in Fig.~\ref{fig:lane_change_fast_speed_Results}.

\begin{figure}[h]
	\begin{centering}
		\includegraphics[scale=0.43]{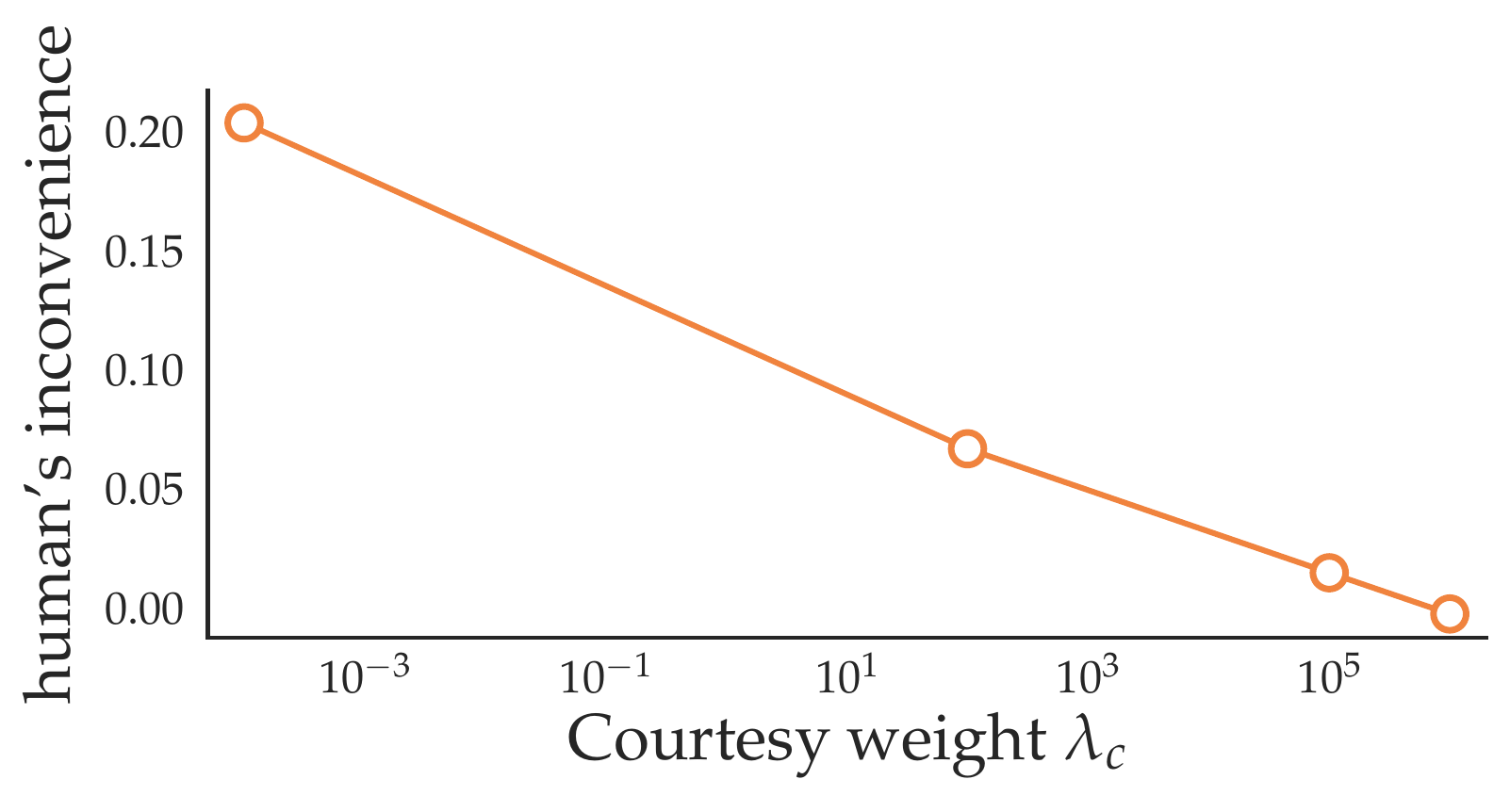}
		\par
	\end{centering}
	\caption{Inconvenience to the human decreases as $\lambda_c$ increases.}\label{fig:sacrifice_curve}
\end{figure}

Figure~\ref{fig:sacrifice_curve} summarizes the relationship between the human driver's inconvenience (the magnitude of the courtesy term) and $\lambda_c$ for the simulation conditions in Fig.~\ref{fig:lane_change_low_speed_Results}. One can note that as the courtesy of the autonomous car increases, the human driver's inconvenience decreases.

\subsubsection{Turning Left}
\begin{figure*}[ht!]
	\begin{centering}
		\includegraphics[scale=0.54]{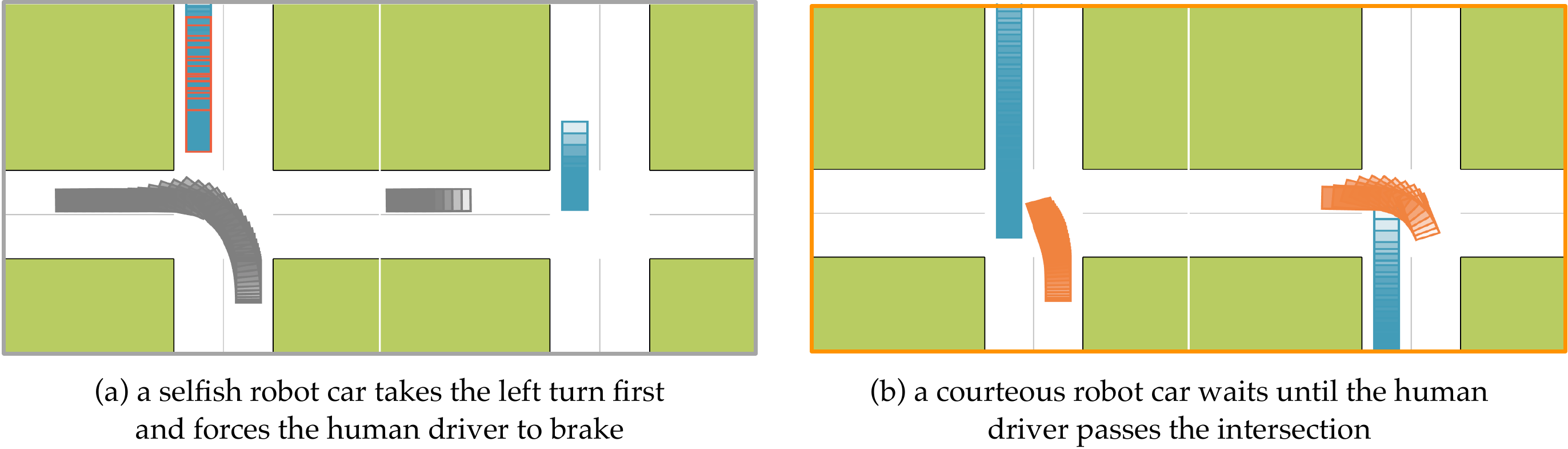}
		\par\end{centering}
	\caption{Interaction between a straight-driving human and a left-turning autonomous car: (a) a selfish (baseline) robot car takes a left turn immediately and forces the human driver to brake (red frames); (b) a courteous robot car waits in the middle of the intersection and takes the left turn after the human passes so that the human can maintain speed.\label{fig:intersection_Results}}
\end{figure*}

In this scenario, an autonomous car wants to take a left turn at an intersection with a straight-driving human. In this case as well, the alternative behaviors that we consider when evaluating inconvenience are the same among three different alternatives: the human driver crosses the intersection maintaining speed. 
 
Simulation results with a courteous and selfish autonomous car are shown in Fig.~\ref{fig:intersection_Results}, where a selfish robot car takes a left turn immediately and forces the human driver to brake (Fig.~\ref{fig:intersection_Results}(a)); while a courteous robot car waits in the middle of the intersection and takes the left turn after the human driver passes the intersection so that the human can maintain its speed (Fig.~\ref{fig:intersection_Results}(b)).

\subsection{Influence of Different Alternative Costs for Evaluating Inconvenience}
In the previous examples, the human would have arrived at the same trajectory regardless of which alternative world we are considering to evaluate how much inconvenience the autonomous car is causing. Here, we consider a scenario in which that is no longer the case to highlight the differences generated by the alternative formulations of courtesy in the robot car's behavior.

We consider a scenario where the human is turning right, with a straight-driving robot car coming from their left. In this scenario, the three alternative costs are different, which leads to different courtesy terms:
\begin{itemize}
	\item Alternative I\---Robot car not being there: the optimal human behavior would be to take a right turning directly;
	\item Alternative II\---Robot car being collaborative: the robot would take the necessary yielding maneuver to let the human driver take the right turn first, leading to the same alternative optimal human behavior of performing the right turn directly;
	\item Alternative III\---Robot car maintaining behavior: the robot car would maintain its speed, and the optimal human behavior would be to slow down.
\end{itemize}

\begin{figure}[ht!]
	\begin{centering}
		\includegraphics[scale=0.6]{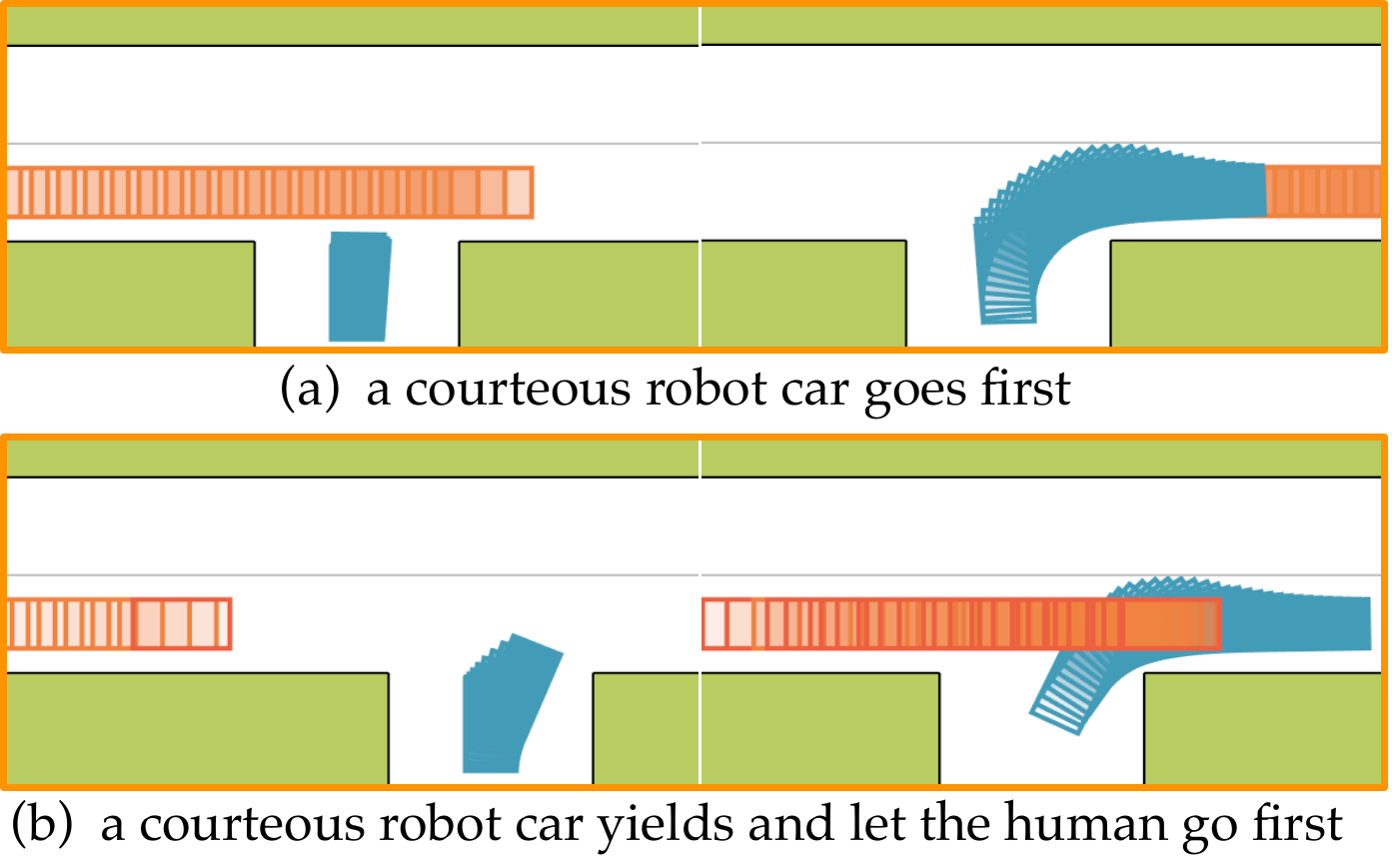}
		\par\end{centering}
	\caption{Interaction between a right-turning human driver and a courteous autonomous car with different courtesy terms: (a) the robot car goes first when it evaluates the courtesy term using going forward as an alternative world; (b) the robot car yields and let the human go first when it evaluates the courtesy term based on a collaborative or not-being-there alternative world.}
	\label{fig:Right_turning_human}
\end{figure}

Figure \ref{fig:Right_turning_human} summarizes the results of using these different courtesy terms. In Alternative III, a courteous robot car goes first, as shown in Fig.~\ref{fig:Right_turning_human}(a). Intuitively, this is because $C^{alt,*}_{\Hu}$ is initially high, and by maintaining its speed (or even accelerating depending on $C^{self}_{\R}$), no further inconvenience is brought to the human by the robot car, i.e., $C^{court}_{\R}$ remains zero. Hence, the robot car goes first (Had the robot try to brake, it only increases $C^{self}_{\R}$ without changing $C^{court}_{\R}{=}0$, and therefore $C_{\R}$ increases). The other two alternatives (I and II) are much more generous to the human. Results in Fig.~\ref{fig:Right_turning_human}(b) show that a courteous robot car finds it too expensive to force the human to go second, and slows down to let the human go first. The red frames in Fig.~\ref{fig:Right_turning_human}(b) indicate the time instants when the autonomous car brakes. 

\subsection{Extension to environments with multiple agents}
\begin{figure}[ht!]
	\begin{centering}
		\includegraphics[scale=0.58]{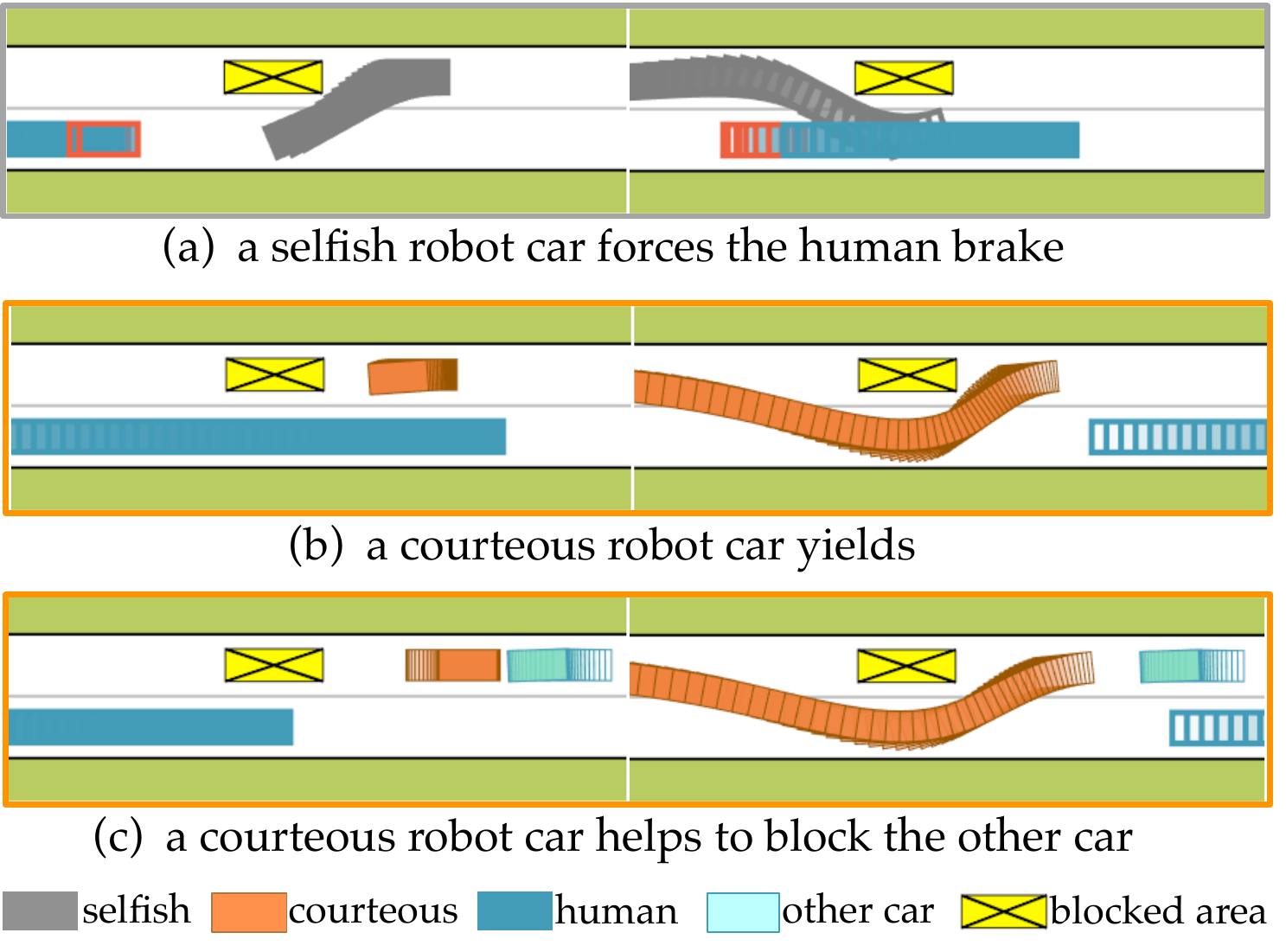}
		\par
	\end{centering}
	\caption{A blocking-area overtaking scenario: (a) with a selfish cost function, the robot car overtakes first and forces the human driver to brake; (b)(c) a courtesy-aware robot car yields to the human driver and even helps to block other cars depending on its formulation of the human driver's alternative world\label{fig:multiple_agent}}
\end{figure}

We study a scenario on a two-way road. The robot car and the human are driving towards opposite directions, but the robot car is blocked and it has to temporarily merge into the human driver's lane to get through, as in Fig.~\ref{fig:multiple_agent}. We use the collaborative robot as our alternative formulation of the courtesy term in this scenario. 

When there are only two agents in the environment, i.e., the autonomous car and the human driver, the results for a selfish and a courteous autonomous car are shown in Fig.~\ref{fig:multiple_agent}(a)-(b): A selfish autonomous car directly merges into the human's lane and forces the human driver to brake; while a courteous autonomous car decides to wait until the human driver passes by since the courtesy term becomes too expensive to go first. 

Such courtesy-aware planning becomes much more interesting when there is a third agent in the environment, as shown in Fig.~\ref{fig:multiple_agent}(c). We assume that the third agent is a responsive agent to the autonomous car and the autonomous car is courteous only to the human driver (and not to both). In this case, for $C^{alt,*}_{\Hu}$, the human would ideally want to pass undisturbed by either the robot or the other agent: the courtesy term captures the difference in cost to the human between the robot's behavior and the alternative of a collaborative robot, and this cost to the human depends on how much progress the human is able to make and how fast. As a result, a very courteous robot has an incentive to produce behavior that is as close as possible to making that happen. 

Then an interesting behavior emerges:  the autonomous car first backs up to block the third agent (the following car) from interrupting the human driver until the human driver safely passes them, and then the robot car finishes its task. This displays truly collaborative behavior, and only happens with high enough weight on the courtesy term. This may not be practical for real on-road driving, but it enables the design of highly courteous robots in some particular scenarios where human have higher priority over all other autonomous agents.

\section{Courtesy Helps Explain Human Driving}\label{sec:IRL}
Thus far, we have shown that courtesy is useful for enabling cars to generate actions that do not cause inconvenience to other drivers. We have also seen that the larger the weight we put on the courtesy term, the more the car behavior becomes social. A natural next question is -- are humans courteous? 

Our hypothesis is that our courtesy term can help explain human driving behavior. If that is the case, this has two important implications: it means that it can enable robots to better predict human actions by giving them a more accurate model of how people drive, and it also means that robot can use courtesy to produce more human-like driving. 

We put our hypothesis to the test by learning a cost function from human driver data, with and without a courtesy feature. We find that using the courtesy feature leads to a more accurate cost function that is better at reproducing human driver data, lending support to our hypothesis.


\subsection{Learning Cost Functions from Human Demonstrations}
\subsubsection{Human Data Collection}
The human data is collected from the Next Generation SIMulation (NGSIM) dataset \cite{alexiadis_next_2004}, which captures the highway driving behaviors/trajectories by digital video cameras mounted on top of surrounding buildings. We selected 153 left-lane-changing driving trajectories on Interstate 80 (near Emeryville, California), and separated them into two sets: a \emph{training} set  of size 100 (denoted by $\mathcal{U}_D$, i.e., the human demonstrations), and the other 53 trajectories as the \emph{test} set.
\subsubsection{Learning Algorithm}
We use Inverse Reinforcement Learning (IRL) \cite{abbeel2004apprenticeship, ziebart2008maximum, abbeel2011inverse, Levine2012ICML} to learn an appropriate cost function from human data. 

We assume that cost function is parameterized as a linear combination of features:
\begin{equation}
c(x^t, u^t_{\R},u^t_{\Hu};\theta) = \theta^T \phi(x^t, u^t_{\R}, u^t_{\Hu}).
\label{eq:selfish_cost}
\end{equation}
Then over the trajectory length $L$, the cumulative cost function becomes
\begin{IEEEeqnarray}{rCl}
C(x^0, {\mathbf{u}}_{\R},{\mathbf{u}}_{\Hu};\theta)&=&\theta^T \sum_{t=0}^{L{-}1}\phi(x^t, u^t_{\R}, u^t_{\Hu})\nonumber\\
&=&\theta^T \Phi(x^0, \mathbf{u}_{\R}, \mathbf{u}_{\Hu})\label{eq:selfish_cost_cumulative}
\end{IEEEeqnarray}
where $\mathbf{u}_{\R}$ and $\mathbf{u}_{\Hu}$ are, respectively, the actions of the robot car and the human over the trajectory. Our goal is to find the weights $\theta$ which maximizes the likelihood of the demonstrations:
\begin{equation}
\theta^*=\arg\max_{\theta}P(\mathcal{U}_{D}|\theta)\label{eq:optimal_lambda}
\end{equation}
Building on the principle of maximum entropy, we assume that trajectories are exponentially more likely when they have lower cost:
\begin{equation}
    {P(\mathbf{u}_{\Hu},\theta)} \propto \exp\left(-C(x^0, \mathbf{u}_{\R},\mathbf{u}_{\Hu};\theta)\right).
\end{equation}
Thus the probability (likelihood) of the demonstration set becomes 
\begin{equation}
P(\mathcal{U}_{D}|\theta)=\Pi_{i=1}^{n}\dfrac{ P(\mathbf{u}^D_{\Hu,i},\theta)}{P(\theta)}=
\Pi_{i=1}^{n}\dfrac{ P(\mathbf{u}^D_{\Hu,i},\theta)}{\int P(\tilde{\mathbf{u}}_{\Hu},\theta)d\tilde{\mathbf{u}}_{\Hu}}\label{eq:maximum_entropy}   
\end{equation}
where $n$ is the number of trajectories in $\mathcal{U}_D$.

To tackle the partition term $\int P(\tilde{\mathbf{u}}_{\Hu},\theta)d\tilde{\mathbf{u}}_{\Hu}$ in (\ref{eq:maximum_entropy}), we approximate $C(x^0, {\mathbf{u}}_{\R},\tilde{\mathbf{u}}_{\Hu};\theta)$ with its Laplace approximation as proposed in \cite{Levine2012ICML}:
\begin{IEEEeqnarray}{rCl}
C(x^0{,}{\mathbf{u}}_{\R},\tilde{\mathbf{u}}_{\Hu};\theta)&\approx& C(x^0, {\mathbf{u}}_{\R},{\mathbf{u}}^D_{\Hu,i};\theta){+}\left(\tilde{\mathbf{u}}_{\Hu}{-}{\mathbf{u}}^D_{\Hu,i}\right)^{T}\dfrac{\partial C}{\partial {\mathbf{u}}_{\Hu}}\nonumber\\
&&+\dfrac{1}{2}\left(\tilde{\mathbf{u}}_{\Hu}{-}{\mathbf{u}}^D_{\Hu,i}\right)^T\dfrac{\partial^2 C}{{\partial} \mathbf{u}^2_{\Hu}}\left(\tilde{\mathbf{u}}_{\Hu}{-}{\mathbf{u}}^D_{\Hu,i}\right).\nonumber\\
\label{eq:laplace_approximation}
\end{IEEEeqnarray}
With the assumption of locally optimal demonstrations, we have $\dfrac{\partial C}{\partial \mathbf{u}_{\Hu}}|_{\mathbf{u}^D_{\Hu,i}}{=}0$ in (\ref{eq:laplace_approximation}). This simplifies the partition term $\int P(\tilde{\mathbf{u}}_{\Hu},\theta)d\tilde{\mathbf{u}}_{\Hu}$ as a Gaussian Integral where a closed-form solution exists (see \cite{Levine2012ICML} for details). Substituting (\ref{eq:maximum_entropy}) and (\ref{eq:laplace_approximation}) into (\ref{eq:optimal_lambda}) yields the optimal parameter $\theta^*$ as the maximizer.

\subsection{Experiment Design} 

\noindent\textbf{Hypothesis.} Within human interactions, human drivers show courtesy to others, i.e., they optimize a compound cost function in the form of $C=C^{self}+\lambda_c C^{court}$ as (\ref{eq:courtesy_cost}) instead of a selfish one as $C^{self}$.

\noindent\textbf{Independent Variable.}
To test our hypothesis, we run two sets of IRL on the same set of human data, but with one different feature. For the selfish cost function $C^{self}$, four features are selected as follows:
\begin{itemize}
	\item speed feature $f_d$: deviation of autonomous car's speed compared to the speed limit: \begin{equation}\label{eq:fd}
        f_d=(v-v_d)^2
    \end{equation}
	\item comfort features $f_{acc}$ and $f_{steer}$: jerk and steering rate of the autonomous car;
	\item goal feature $f_g$: distance to the target lane:
	\begin{equation}\label{eq:fg}
        f_g=e^{\frac{d_g}{w_l}},
    \end{equation}
    where $d_g$ is the Euclidean distance and $w_l$ is the lane width.
	\item safety feature $f_s$: relative positions with respect to surrounding cars;
	\begin{equation}\label{eq:fs}
        f_s=\sum_{i=1}^{n_s} e^{-d_i},
    \end{equation}
    where $n_s$ is the number of surrounding cars and $d_i,i{=}0,1,\cdots,n_s$ is the distance to each of them.
\end{itemize}

For the courtesy-aware cost function $C=C^{self}+\lambda_c C^{court}$, we use the same four features as above, plus one additional feature that equals to the courtesy term.

\noindent\textbf{Dependent Measures.}
We measured the similarity between trajectories planned with the learned cost functions and human driving trajectories on the \emph{test} set (another  53 left-lane changing scenarios that are different from the training set from the NGSIM dataset). 

\subsection{Analysis}
\begin{figure}[thpb]
	\centering
	\includegraphics[height=.25\textwidth]{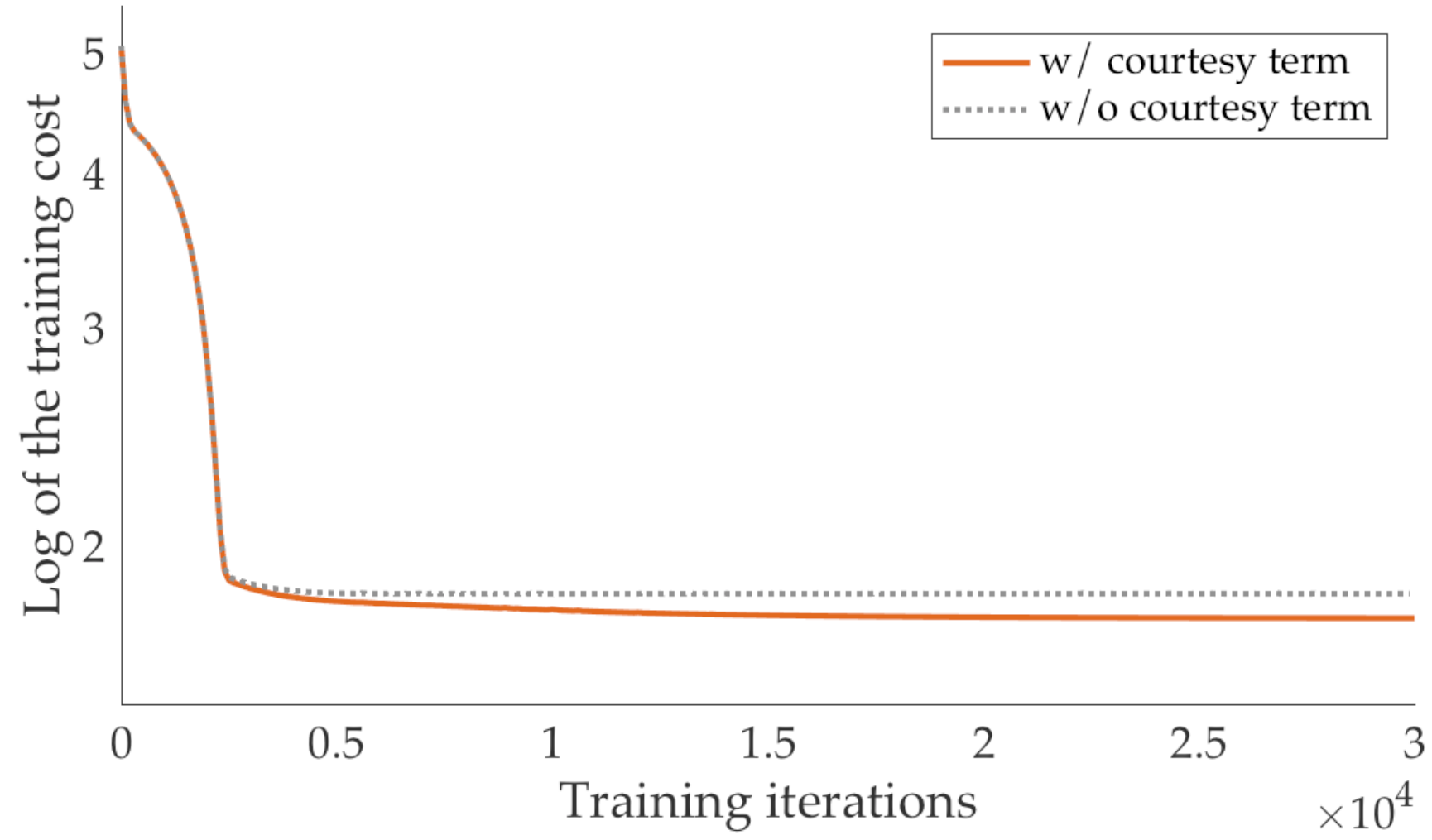}
	\caption{Training curves for cost functions with and without the courtesy term}
	\label{fig:training_curve}
\end{figure}

\noindent\textbf{Training performance.}
The training results are shown in Fig.~\ref{fig:training_curve} and Table \ref{tab:learned_parameters}.
One can see that with the additional courtesy term, better learning performance (in terms of training loss) has been achieved. This is a sanity check: having access to one extra DOF can lead to better training loss regardless, but if it did not that would invalidate our hypothesis.  
\begin{table}[th!]
    \centering
    \begin{tabular}{|c|c|c|c|c|c|c|}
    	\hline 
    	& $\theta_g$ & $\theta_d$ & $\theta_{acc}$ & $\theta_{steer}$ & $\theta_{s}$ & $\lambda_c$ \tabularnewline
    	\hline 
    	\hline 
    	$C^{self}$ & 1.0 & 2.08e+04 & 5.80e+02 & 3.91e+02 & 4.37 & -- \tabularnewline
    	\hline 
    	$C$ & 1.0 & 1.96e+02 & 6.7e+04 & 2.36e+02 & 6.53 & 9.89e+04 \tabularnewline
    	\hline 
    \end{tabular}
    \caption{The parameters in $C$ learned via IRL}
    \label{tab:learned_parameters}
\end{table}

\begin{figure}[thpb]
	\centering
	\includegraphics[scale=0.43]{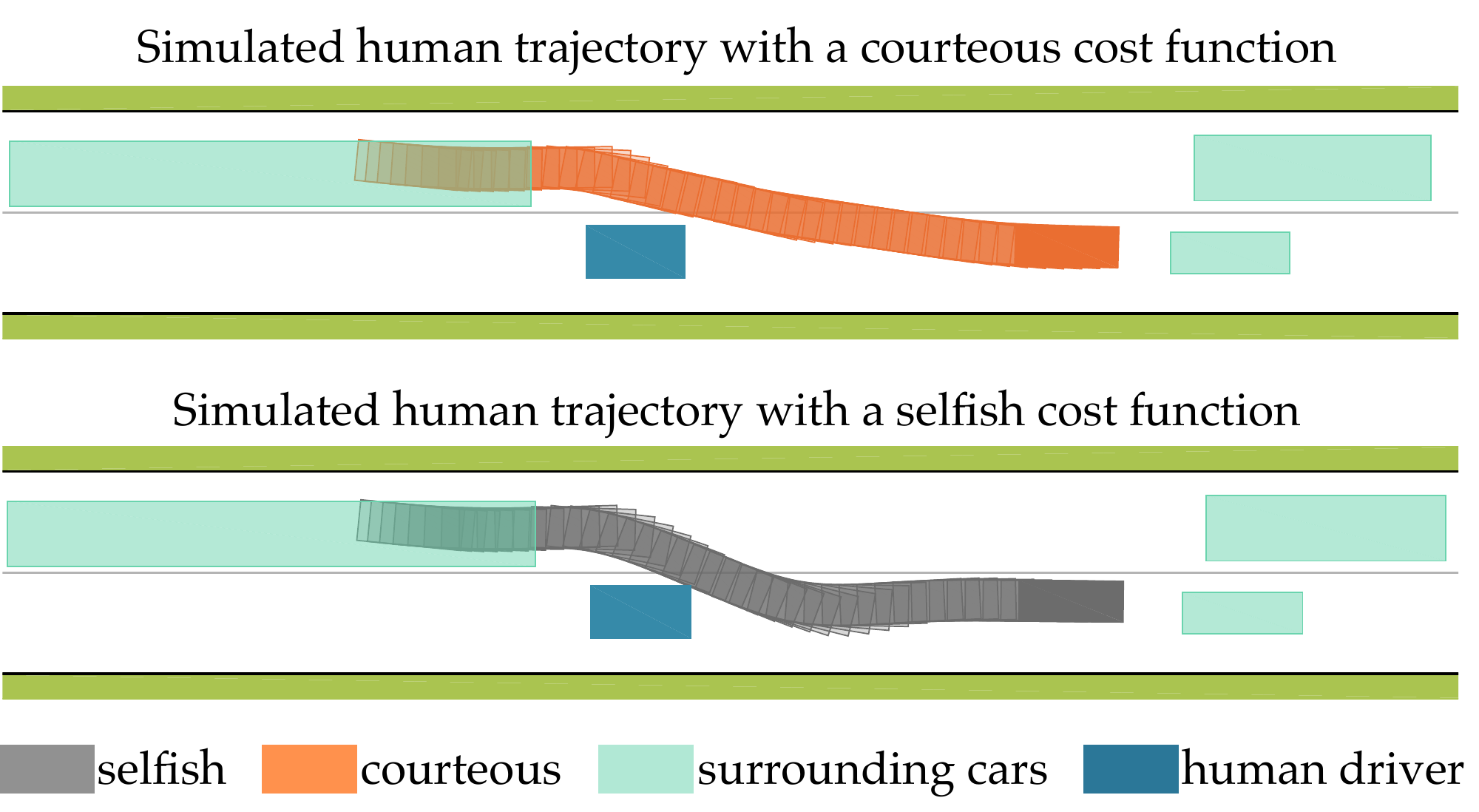}
	\caption{An example pair of simulated trajectories with courteous (top) and selfish (bottom) cost functions}
	\label{fig:trj_example_courtesy}
\end{figure}

\noindent\textbf{Trajectory similarity.}
Figure \ref{fig:trj_example_courtesy} shows one demonstrative example of the trajectories for a selfish car (grey) and a courteous car (orange), with four surrounding vehicles. The dark blue rectangle is the human driver in our two-agent robot-human interaction system and all other vehicles (cyan) are treated as moving obstacles. It shows that a simulated car with $C$ that includes courtesy manages to reduce its influence on the human driver by choosing a much smoother and less aggressive merging curve, while a car driven by $C^{self}$ merges in much aggressively. 

\begin{figure*}[t!] \centering
\includegraphics[height=.21\textwidth]{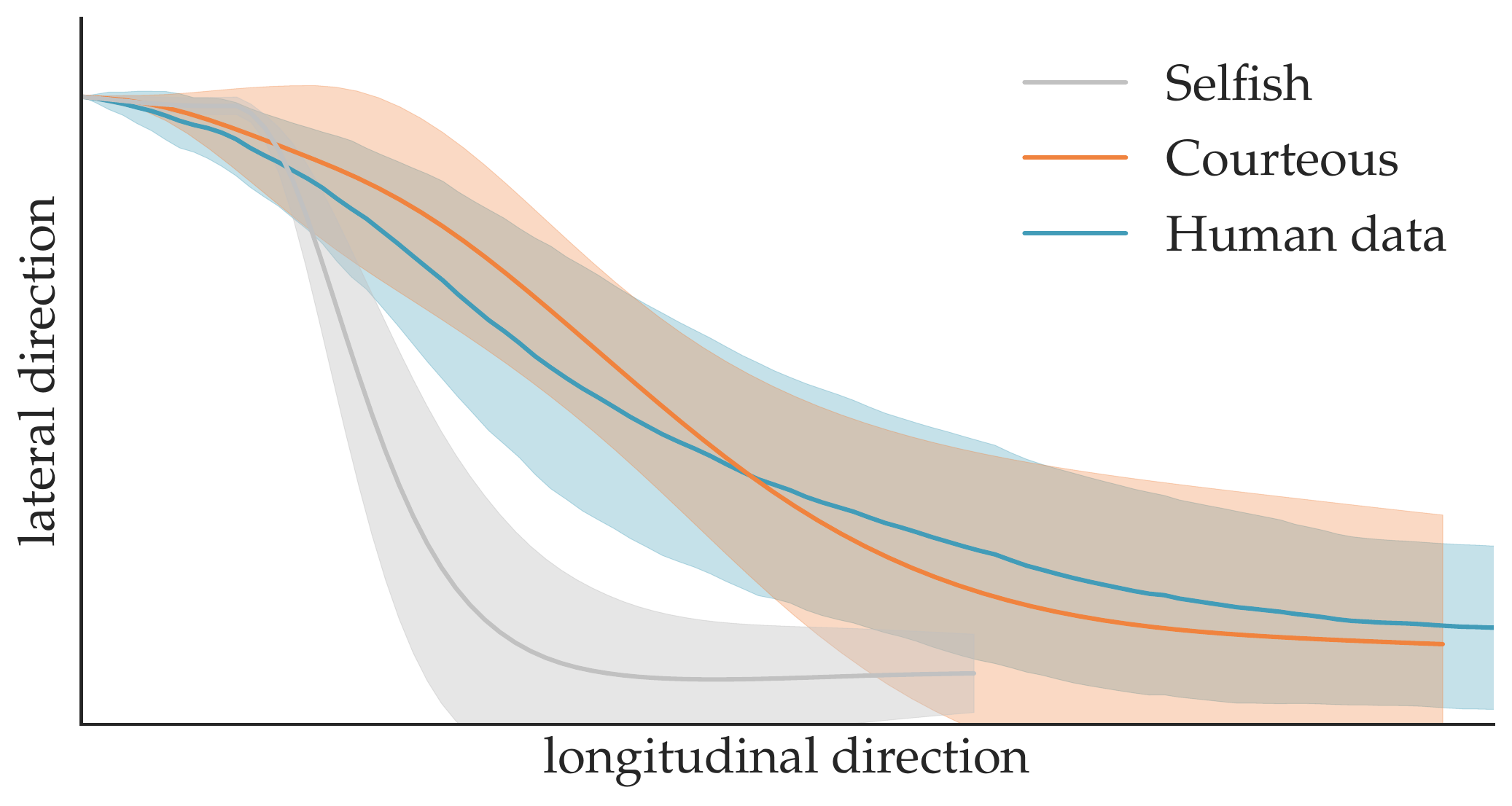}
\includegraphics[height=.21\textwidth]{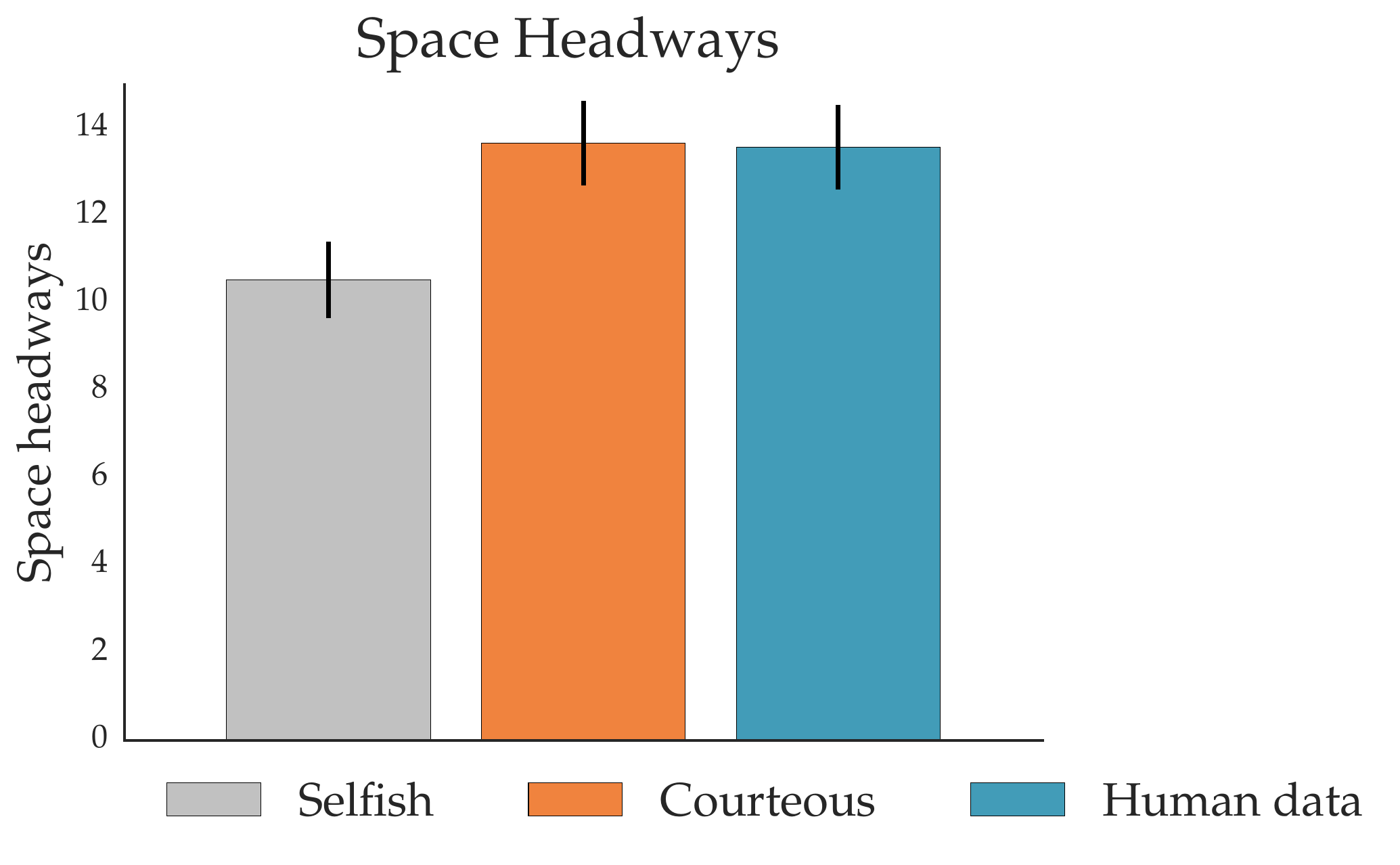}
\includegraphics[height=.21\textwidth]{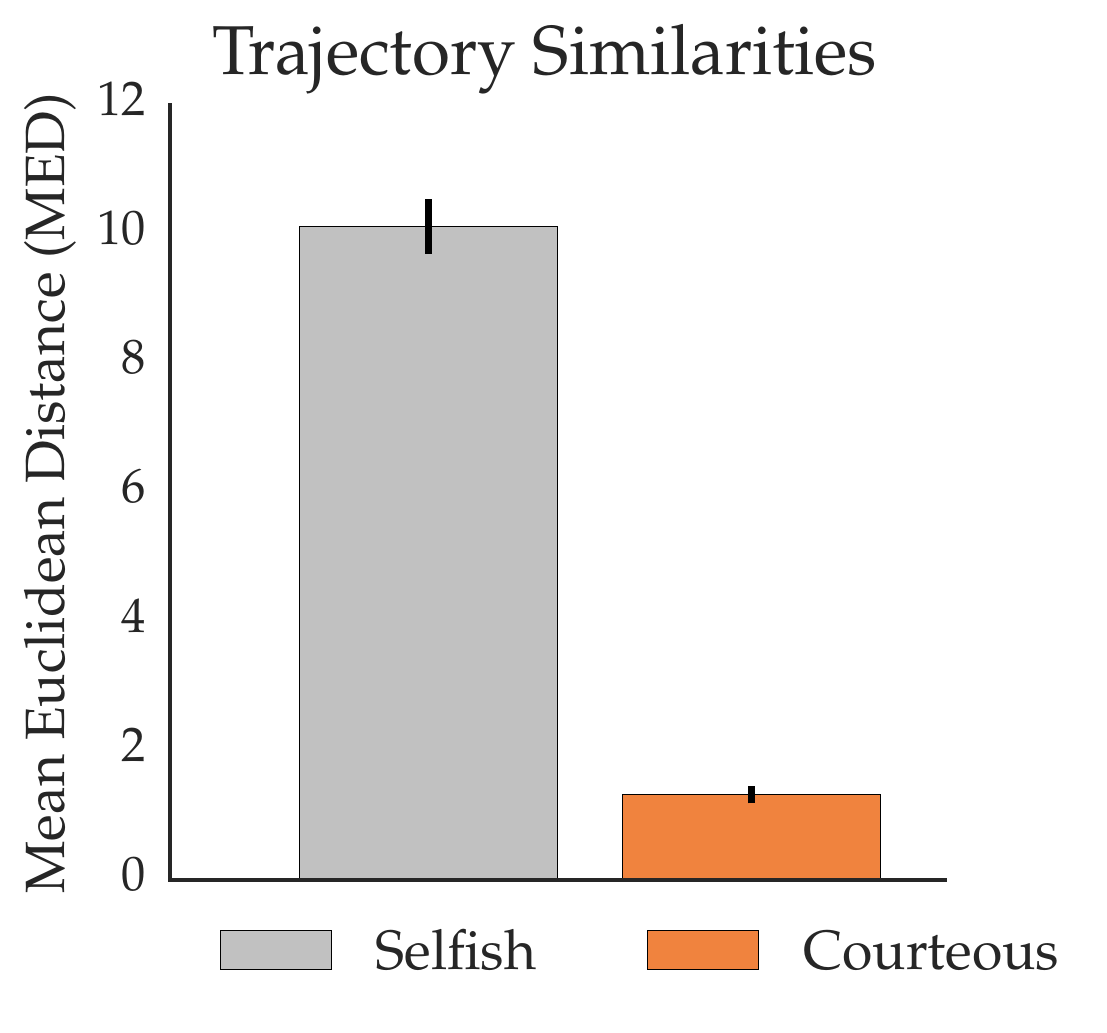}
\caption{The courtesy term helps fit test set human driver data significantly better: we can see this from the actual trajectories (left), the following gaps (middle), and the mean euclidean distances from the ground truth human data (right).}
\label{fig:similarity}
\end{figure*}

 Results for all 53 left-lane changing test trajectories are given in Fig.~\ref{fig:similarity} (left). To describe the similarities among trajectories, we adopted the Mean Euclidean Distance (MED) \cite{quehl_how_2017}. As shown in Fig.~\ref{fig:similarity} (right), the courtesy-aware trajectories are much similar to the ground truth trajectories, i.e., a courteous robot car behaves more human-like. We have also calculated the space headways of the following human driver on the robot car's target lane for all 53 test scenarios, and the statistical results are given in Fig.~\ref{fig:similarity} (middle). Compared to a selfish robot car, a courteous robot car can achieve safer left-lane changing behaviours in terms of following gaps for the human driver behind.

\section{Conclusion}
\label{sec:conc}
\noindent\textbf{Summary.}
We introduced courteous planning based on the fact that human irrationally care more about additional inconvenience they are brought to by others. Courteous planning enables an autonomous car to take into consideration such inconvenience when evaluating its possible plans. We saw that not only this leads to more courteous robot behavior, but it also helps explain real human driving data, because humans too are likely trying to be courteous.

\noindent\textbf{Limitations and Future Work.} 
Despite the fact that courtesy is not absolute, but relative to how well off the human driver could be, the trade-off between courtesy and selfishness remains a meta-parameter that is difficult to set. In general, defining the right trade-off parameters in the objective function for autonomous cars and robots more broadly remains a challenge. With autonomous cars, this is made worse by the fact that it is not neccessarily a good idea to rely on Inverse Reinforcement Learning\---this might give us models of human drivers, as it did in our last experiment, but that might not be what we want the car to optimize for.

Further, we studied courtesy with a single human driver to be courteous toward (we had other agents, but the robot did not attempt courtesy toward them). In real life, there will be many people on the road, and it becomes difficult to be courteous to all. To some extent, this is alleviated by our definition of courtesy: it is not maximizing everyone's utility, but it is minimizing the inconvenience we cause. But further work needs to push courtesy to the limits of interacting with multiple people in cases where it is difficult to be courteous to all. 

\section*{Acknowledgement}
This work was partially supported by Mines ParisTech Foundation, “Automated Vehciles\---Drive for All” Chair, and NSF CAREER. We thank Jaime F. Fisac for helpful discussion and feedback.



\addtolength{\textheight}{-10cm}   


\bibliographystyle{IEEEtran}
\bibliography{IROS2018_Courtesy}
\end{document}